\documentclass[conference]{IEEEtran}
\IEEEoverridecommandlockouts
\usepackage{biblatex}
\usepackage{amsmath,amssymb,amsfonts}
\usepackage{algorithmic}
\usepackage{graphicx}
\usepackage{svg}
\usepackage{amsmath}
\usepackage{array}
\newcolumntype{L}{>{\centering\arraybackslash}m{3cm}}
\usepackage{textcomp}
\usepackage[raggedrightboxes]{ragged2e}
\usepackage{xcolor}
\begin{document}

\title{Guidance system for Visually Impaired Persons using Deep Learning and Optical flow\\
}

\author{\IEEEauthorblockN{1\textsuperscript{1st} Shwetang Dubey}
\IEEEauthorblockA{\textit{(Member IEEE)}\\{Information Technology} \\
\textit{Indian Institute of information technology, Allahabad}\\
Prayagraj, India \\
pro2017002@iiita.ac.in}
\and
\IEEEauthorblockN{2\textsuperscript{2nd} Alok Ranjan Sahoo}

\IEEEauthorblockA{\textit{(Member IEEE)}\\{Information Technology} \\
\textit{Indian Institute of information technology, Allahabad}\\
Prayagraj, India \\
rsi2017503@iiita.ac.in}
\and
\IEEEauthorblockN{3\textsuperscript{3rd} Pavan Chakraborty}
\IEEEauthorblockA{\textit{(Member IEEE)}\\{Information Technology} \\
\textit{Indian Institute of information technology, Allahabad}\\
Prayagraj, India \\
pavan@iiita.ac.in}
% \and
% \and
% \and
 }

\IEEEpubid{\makebox[\columnwidth]{979-8-3503-4210-9/23/\$31.00~\copyright2023 IEEE \hfill} \hspace{\columnsep}\makebox[\columnwidth]{ }}

\maketitle

\begin{abstract}

Visually impaired persons find it difficult to know about their surroundings while walking on a road. Walking sticks used by them can only give them information about the obstacles in the stick's proximity. Moreover, it is mostly effective in static or very slow-paced environments. Hence, this paper introduces a method to guide them in a busy street. To create such a system it is very important to know about the approaching object and its direction of approach. To achieve this objective we created a method in which the image frame received from the video is divided into three parts i.e. center, left, and right to know the direction of approach of the approaching object. Object detection is done using YOLOv3. Lucas Kanade's optical flow estimation method is used for the optical flow estimation and Depth-net is used for depth estimation. Using the depth information, object motion trajectory, and object category information, the model provides necessary information/warning to the person. This model has been tested in the real world to show its effectiveness.

\end{abstract}


\section{Introduction}
According to a report published by WHO on 14 October 2021\cite{WHO}, there are at least 2.2 billion peoples who are having vision impairment. Around 1 billion are having moderate or severe vision impairment or blindness. The number was around 285 million in 2010. 246 million were having serious blindness. So, the rise in visual impairment necessitates the development of an algorithm to give real-time suggestions to fully or partially blind persons. Hence, this paper intends to give a method for guiding visually impaired persons in real to avoid obstacles. 

\hspace{10mm} The real challenge with the guidance system is the response time of that model. As when a person is moving on the road then it means that when any object comes in front of him then he will not be able to avoid it unless the guidance system gives him a good response in real-time. So the main challenge here was to generate a system that can not only give good object detection and depth estimation but also can give better time accuracy so that obstacle avoidance can be performed in real-time. 

\hspace{10mm} Here we are using the YOLOv3\cite{yolov3} model for object detection as it can give out output much faster than other object detection algorithms like R-CNN, Fast R-CNN, and Mask R-CNN \cite{Fast_RCNN}\cite{Mask_RCNN}\cite{R_CNN}. Since all YOLO models are trained to do classification and bounding box regression simultaneously so it works much faster than R-CNN or Fast R-CNN. This YOLOv3 object detection algorithm is trained on the COCO\cite{COCO} dataset which has a total of 80 classes. So we will be able to detect any object if they belong to one of these 80 classes. This YOLOv3 algorithm generates an Anchor box around the detected object which can be further used for depth estimation if we have the feed from both the left and right camera. 

\hspace{10mm} In this paper we are giving a model to generate voice commands for the user in a real-time situation for assisting a visually impaired person to move. We also performed optical flow analysis to get the motion of the approaching object towards the camera to generate proper instruction for the blind person. We have divided the frame into 3 parts i.e. left, right, and center then according to the position of the approaching object, the instruction is generated. The main contribution of this paper is the following:
\begin{itemize}
    \item Used YOLOv3 object detection algorithm for classification of objects in video frames from both left and right images.
    \item A method to divide the entire image into 3 segments i.e. left, right, and center to generate instruction properly.
    \item Created an optical flow-based method to give the trajectory of the moving object for object tracking.
    \item Depth estimation is also performed for detecting depth in real time.
    \item Finally Google text-to-speech converter is used for generating speech instruction for blind persons.

\end{itemize}
\section{RELATED WORK}
In recent years lots of methods have been developed for vision-based obstacle detection and avoidance systems but most of them use different technologies like WiFi, RFID, laser, etc. but the use of cameras for object detection and avoidance is very limited. Vision-based obstacle avoidance system was first introduced by Sainarayanan et al.\cite{Sainarayanan}. In this the team has used they used grayscale images for detection and the background removal is done by using a neural network and obstacle pixels are enhanced. Ulrich et al.\cite{ulrich} proposed a method of histogram comparison. In this method, first color image is filtered and then converted to HSV color space and then the color histograms of the candidate area and reference area are matched. Joachim et el.\cite{Joachim} derived a method to detect obstacles using human color vision and then the auto-focused stereo camera is used to get the depth of obstacles from a person. Rodríguez et al.\cite{Rodriguez} proposed a model which was based on the cumulative grid in front of visually impaired users for obstacle detection and avoidance. In this method basically a stereo vision is used for obstacle detection and background reduction. 
\subsection{Obstacle detection}
\hspace{10mm} Bernabei et el.\cite{Bernabei} proposed a method that basically uses an RGB-D sensor for depth estimation and obstacle detection. In this paper basically, a 3D points cloud is generated using Microsoft Kinect, and then the volume of the obstacle in front of the person is calculated finally according to the volume instructions are generated. Vlaminck et al.\cite{Vlaminck} presented a method where he used RANSAC on a 3D point cloud for plane segmentation. After getting different planes they used that information for ground, wall, and obstacle detection. This method is good but not in real-time situations as RANSAC takes too much time to process the 3D point cloud. Moreover, in this paper, they considered obstacles to be on ground level which is not true for all cases. Rodríguez et al.\cite{Rodriguez} published a paper that uses a stereo camera for screen capture and then they generated a map of that place using visual SLAM and used that map for autonomous navigation of the visually impaired person. 

\subsection{Feedback system using voice command}
\hspace{10mm} After detecting the obstacle the second part of this problem is to make an alert mechanism that can alert the visually impaired person in real-time. This is a hard task to describe where the location of an obstacle and how to avoid it to a blind person as he can not understand the world as we do. There are multiple ways used to describe scenes by different researchers. Joachim et al.\cite{Joachim} used a text-to-speech engine to convert the text into the form of speech and then used a speaker to alert a person.  The vOICe \cite{voice} system is introduced to give complete views of the scene through image-to-sound renderings. In this system the image is scanned from left the right and any elevation is represented as pitch and brightness as loudness. So using sound illusion created by the speaker a rough figure of the scene is created in the person's mind. This system is still in the development stage and needs a lot of practice from users to clearly understand how to use it. This is why this system is usually suitable for younger people. Sainarayanan et al. \cite{Sainarayanan} used the segmentation of the image into two parts left and right for this purpose. The warning voice is generated according to the position of the obstacle in these two segments and then sent to the user through headphones. 
\subsection{Feedback system using tactical sensors}
\hspace{10mm} Johnson and Higgins et al.\cite{jhonson} created a tactical feedback system in which different motors are attached to generate the vibration signal for the obstacle present in front of the user. According to the location of the obstacle, motors are assigned. This system needs training of the user for proper functioning. Nguyen et al.\cite{Nguyen} used an eclectic pulse-based system for feedback. Electric pulses are generated in data gloves and give instructions through nerves present in the skin.

\section{Method}
In this section, we will describe how we performed our experiment and what are different methods used for obstacle detection, depth estimation, and warning generation. We are basically using the object detection method for obstacle detection in both stereo inputs and then depth is estimated using stereo vision and finally warning is generated using a GOOGLE text-to-speech generation algorithm.

\subsection{Obstacle detection using YOLOv3}
\hspace{10mm}Obstacle detection is one of the basic problems to create effective methods for visually impaired guidance systems. As we know obstacles can be on any elevation and will keep moving towards the person or going away from the person so the location of an obstacle is keep changing in a real-time environment. To detect those objects and classify them as obstacles is one of the biggest challenges of this problem. Here we are using the YOLOv3\cite{yolov3} object detection algorithm for obstacle detection because it is very fast and creates anchor boxes around the object. YOLOv3 is trained on the COCO dataset which has 80 classes so it can predict the object if it belongs to one of these classes. YOLOv3 uses Darknet-53 as a backbone feature  extractor which has 53 convolutional layers making it a powerful network. YOLOv3 has skip connection-based architecture like ResNet and 3 prediction heads like FPN. Basically, YOLOv3 is not as accurate as YOLOv4\cite{yolov4} or YOLOv5 but this is much faster so we used this model for our method as it can give results in real-time, and using that we can create a good guidance system. We have used YOLOv3 in our captured frames for obstacle detection and then monocular depth estimation is performed for the depth estimation of the object.

\begin{figure}[htbp]
\centerline{\includegraphics[width=9cm]{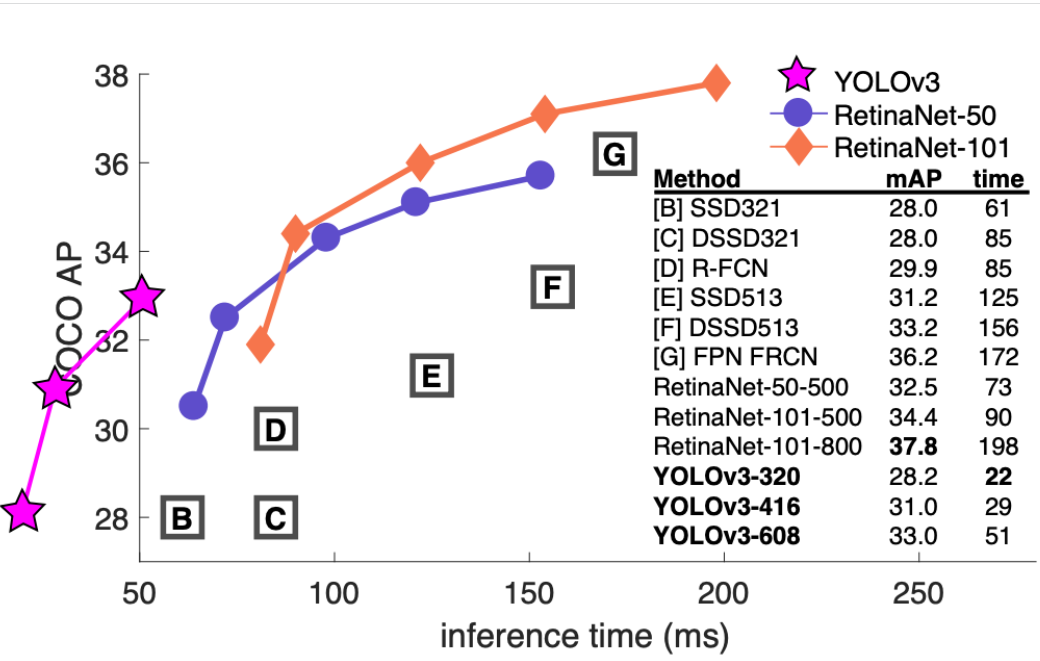}}
\caption{YOLOv3 comparison with other object detection models. This shows YOLOv3 is faster and better this is why we are using it for our model.\cite{yolov3}}
\end{figure}

\subsection{Depth estimation}
\hspace{10mm}Depth estimation is one of the basic needs for this type of problem as we need to find the depth of the obstacle from the person to generate a warning in time. Since we are using only one camera for our objective so depth estimation becomes one of the most challenging works for this problem. Monocular depth estimation is often considered an ill-posted problem as estimating depth only from the pixel values is not generally possible. But current development in deep learning techniques made it possible up to a good extent. There are many deep learning-based techniques available for depth estimation like FlowNet architecture by Dosovitskiy et al.\cite{flownet}, who applied a supervised encoder-decoder CNN-based method to estimate the optical flow using channel-concatenated image pairs, and Zhou et al.\cite{zhou} who used unsupervised settings for depth and pose estimation from a video sequence.

\begin{figure}[htbp]

\centerline{\includegraphics[width=9cm]{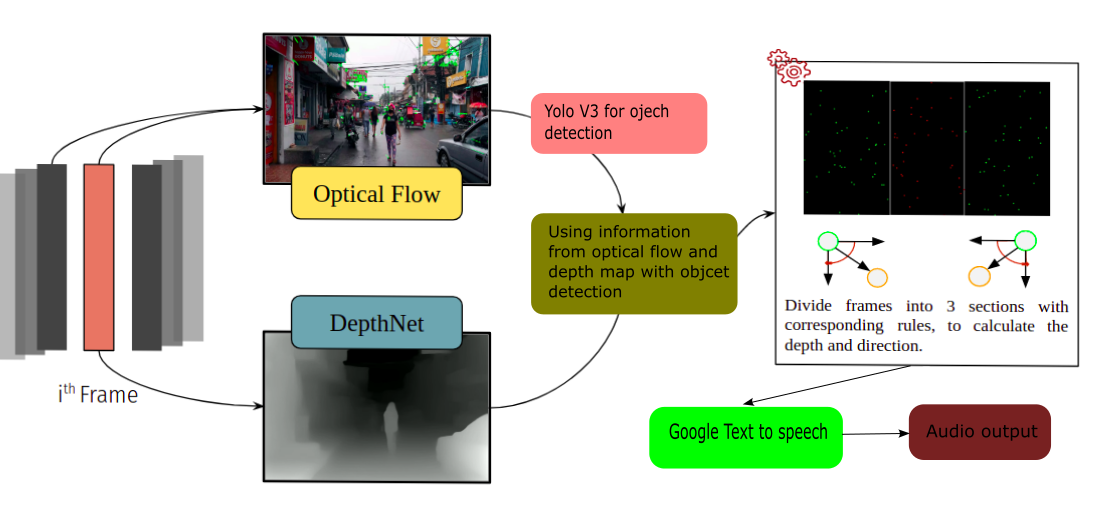}}  
  \caption{The $I^th$ frame in the figure is a current frame and optical flow uses current and previous frames for flow detection. YOLOv3 model is used for object detection and Depth net is used for depth prediction from $I^th$ frame. Finally, all three inputs are given to the model to generate a warning signal.}
\end{figure}

In our approach, we used DephtNet\cite{depth_net} for depth estimation which is a recurrent neural network architecture for monocular depth estimation. In DepthNet convolutional LSTM (ConvLSTM)- based architecture is used for depth estimation and the fully connected LSTM layer is replaced by a stack of ConvLSTM layers. The LSTM layers allow the network to learn temporal information better but here convolutional layer is also used as it retains the spatial relationship between the grid cells. DepthNet uses multiple frames from the video feed for depth estimation from a scene which goes into an encoder-decoder setup. As this network is trained on the KITTI dataset so it is suitable for outdoor performance tasks. Depth calculated from this method is fast and more accurate as it uses multiple frames of video of the same scene so we are using this method in our task. The loss function used here is the Eigen scale invariant loss function by Eigen et al.\cite{Eigen_loss}. Given a predicted depth map $y_i$ and ground true depth map as $y_i^*$ the loss function is given by:
\begin{equation}
 L(y,y^*) = \frac{1}{n}\sum_i d_i^2 - \frac{\lambda}{n^2} (\sum_i d_i)^2 
\end{equation}
where $d_i = log(y_i) - log(y_i^*)$ for the $i^th$ pixel and $n$ is total number of pixels.

The depth map evaluated from the DepthNet is mixed with the optical flow obtained and then given to the model to generate the warning according to distance from the person. The warning is generated according to the quadrant of the obstacle in the image. The next section is about how we obtained optical flow from the video feed.
\subsection{Optical flow estimation} 
\hspace{10mm} Optical flow or optical flow is the pattern of apparent motion of objects, edges, and surfaces in a visual scene caused by the relative motion between an observer (an eye or a camera) and the scene.
Motion field can be defined as the real-world 3D motion and the optical flow field is its projection to the 2D image. So any motion in 3D generates a vector in 2D and that vector is defined as the optical flow of that particular object. J.L. Barron and N. A Thacker \cite{Barron} proposed that a 2D Motion Field can be defined as -2D velocities for all visible points. We used Lucas-Kanade\cite{lucas} method for optical flow estimation from the video feed. This method has two assumptions for optical flow estimation i.e the two images are separated by a small time increment ${\Delta}t$, in such a way that objects have not displaced significantly and the images depict a natural scene containing textured objects exhibiting shades of gray which changes smoothly.

Since we are using the camera on a person who is moving at a slow speed and also obstacles approaching him are also not that fast so we are satisfying both assumptions up to a good level so we are using this model for our optical flow estimation. Lucas Kanade's\cite{lucas} method solves basic optical flow equations for all the pixels in that neighborhood using least squares error. This method is very fast in operation so it is most suitable for our task as we want to create a real-time system. The disadvantage of this method is errors in moving boundaries. If $P1,P2....Pn$ are pixels of the window and $I_x(p_n),I_y(p_n),I_t(p_n)$ are partial derivatives of the image I with respect to position x,y, and the time t of point $P_n$ at the current time. $V_x and V_y$ are components velocity vector then the equations according to Lucas Kanade's method can be given as:
\[Av = b\], where
\[A = \begin{bmatrix}
I_x(p_1) & I_y(p_1) \\
I_x(p_2) & I_y(p_2) \\
\vdots & \vdots \\
I_x(p_n) & I_y(p_n)

\end{bmatrix}, v = \begin{bmatrix}
V_x \\ V_y
\end{bmatrix}, b = \begin{bmatrix}
-I_t(p_1)\\
-I_t(p_2)\\
\vdots\\
-I_t(p_n)
\end{bmatrix}
\]
Lucas Kanade used a least squares criterion-based approach for optical flow estimation. So: 
\[V = (A^TA)^{-1}A^Tb\]
where, $A^T$ is transpose of matrix $A$,

This equation gives the same importance to all the pixels of the image but we want to give more importance to the center pixels so in the Lucas Kanade method weighted version of the least square is used.
\begin{equation}
    V = (A^TWA)^{-1}A^TWb
\end{equation}
Where W is $n\times n$ dimension diagonal matrix containing weights assigned to pixels $p_n$.

The flow-chart for Lucas Kanade's method can be given as:
\begin{itemize}
    \item Calculate flow $u(i-1), v(i-1)$ from level i-1.
    \item Up sample $u(i-1), v(i-1)$ to create $u^*(i), v^*(i)$ of twice resolution as of level i and then multiply them with 2.
    \item Now as $u^*(i), v^*(i)$ apparent velocities so we apply this block displacement.
    \item Then apply Lucas Kanade to find $u^{'}(i)$ and$ v^{'}(i)$.
    \item $u(i)= u^{*}(i) + u^{'}(i)$\\ $v(i) = v^{*}(i) + v^{'}(i)$\\ Finally we apply these two formulae to find optical flow for $I th$ level.
\end{itemize}

We compute above defined Lucas Kanade to the highest level to get optical flow. So after calculating the optical flow and depth map, we have depth information as well as information about the flow of objects in front of a person so now we will use these two pieces of information in our model to create a model which can be used for real-time prediction of warning for a visually impaired person. 

\subsection{Evaluation of Obstacle Location}
One of the most challenging tasks for generating a guidance system is to find the location of the obstacle in the image or in the video frames. For this particular task, we have used a special kind of method in which we divided the entire image or frame into 3 parts ie. Left, Center, and Right. The left part is basically the left of the person who is having the camera and so as others. Here as we already calculated the depth map using DepthNet\cite{depth_net} and optical flow using Lucas Kanade\cite{lucas} so we created an algorithm to generate an alert using all these three values. As if the object is detected using YOLOv3\cite{yolov3} and it belongs to the right side of the image then it is most likely not going to be an obstacle for the moving visually impaired person but if its depth map and optical flow show it is moving towards the person then warning should be generated. And same goes for left-side obstacles also. So here we generated an algorithm that uses all three inputs i.e. depth map, optical flow, and object detection outputs, and then generates the warning using these three values in real-time. 

We have divided the image into 3 parts, consider if the complete width of the image is 1280 then the image is divided using the following formula $ 0\: to\: 448$ width is considered as left and if the depth value corresponding to those value is less than 210 then a warning is generated that object is approaching from the left. If the width value is between $448\: to\: 832$ and the depth is less than 220 the warning is generated that the obstacle is in front and if the width value is between $832\: to\: 1280$ and depth is less than 210 then a warning is generated that object is approaching from the right. So using this small method we are generating warning text in our program and we are doing it in a real-time environment.
\subsection{Text to speech} 
We are using Google text-to-speech converter API to convert the generated warning text into speech a blind person can listen to. The API used in this method is commonly known as gTTS \cite{gtts}. We used this API because it generates mp3 files for speech and this gTTS supports $30+$ languages like Hindi, English, Tamil, German, etc., and 100+ voices with localized accents for some languages like English, French, Mandarin, etc. so this a very useful tool provided by google for this particular task. Google Text to Speech accepts a maximum of 100 and if the length is longer than that then it is handled by dividing the text into two parts. In our use case, we do not need to do that as we are generating only small text files as warning symbols. The text messages which are used as warnings are:
\begin{itemize}
    \item The object is approaching from the left.
    \item The object is approaching from the center.
    \item The object is approaching from the right.
\end{itemize}

\section{Experiments and Results}
In this section, we will apply our model to the real-time environment and will see how the model performs. We are using a single-camera framework for this purpose where we record the moving scene in front of the visually impaired person. The model is working in real time but for evaluation purposes, we recorded the video and applied the model to our computer. 
\subsection{Datasets} Here we are using the Common objects in Context COCO\cite{COCO} dataset for obstacle detection purposes. This dataset is made of a total of 328K images of 80 different object categories. This dataset was developed by Microsoft for different tasks like object detection, captioning, keypoint detection, etc. This dataset is having images of very common objects like persons, cars, chairs, etc. The video is captured from a single camera installed on top of a moving person. This video is used as input to our model and for frame capturing.
\subsection{Implementation details} In our experiment we used an MP4 video file as input for implementing our model. The output video file is of shape 2560*720 in which the object is detected and written as well as audio based warning is generated. We used Lucas Kanade's method for optical flow estimation, using trajectories made by optical flow to judge from which side the object is approaching the person. We used the search window size for Lucas Kanade as $15\times 15$, maximal pyramid level number used is 4 so a total of five levels have been used. The shi-Tomasi corner detection method is used for corner detection with a maximum number of corners of $200$ and quality level parameter of $0.03$ and a maximum distance of $10$. Depth net used for depth estimation uses encoding and decoding phases where the encoding phase consists of $3 \times 3$ ConvLSTM layers of size $\{32, 64, 64, 128, 128, 256, 256, 512\}$ filters respectively, except first two layers where $7 \times 7$ and $5 \times 5$ filter size is used. A ReLu activation function is used for the ConvLSTM layer and a hard sigmoid is used for the recurrent step. In the decoding phase, a number of deconvolution layers have been used of sizes $\{512,256,256,128,128,64,64,32\}$ respectively. The filter size used for deconvolution layers is $3\times 3$ and $1\times 1$. Finally, we have used a Google text-to-speech converter to convert the text available into a voice signal so that the person can be alerted. We are English language for our experiment since it is mostly accepted language.

\begin{figure}[htbp]
\centerline{\includegraphics[width=9cm]{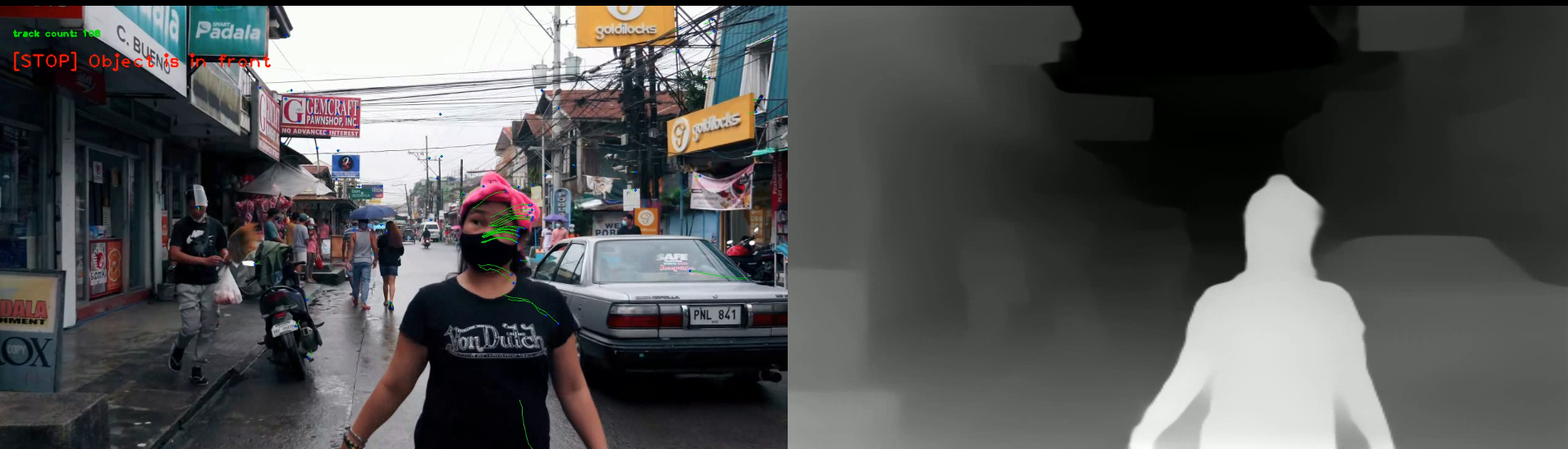}}
\caption{Output of model when a person is coming from the front. The lines are showing the optical flow.}

\end{figure}

\subsection{Experimental setup}
The proposed model uses a Pytorch and Google-Colaboratory GPU for processing purposes. We used a single video camera on the forehead of the moving person to record a video feed for testing. Then we deploy our model to a laptop attached to the forehead camera to generate the warning according to the approaching obstacle. The text-to-speech converter is then used to convert the text which we got from our model to convert into a voice command to instruct the visually impaired person about the incoming obstacles. For now, we also used some random moving person video from the internet to test our model and it is working well in those inputs and giving sufficient output.
\subsection{Testing on videos} 
We have tested this model on different videos available on YouTube for prediction. Using this model the output achieved is fast as the objects can be faster recognized using the YOLOv3 method. As we can see in the Fig.1 the YOLOv3 is faster compared to previous methods in the evaluation part. At the table, we are watching a video with our eyes and observing from which side the person or obstacle is approaching and then we have tested our model on the same video to generate warnings for the visually impaired person. We have repeated this process for three videos and the results are shown in the table. The optical flow model generates warnings only for approaching objects.

\begin{table}[htbp]
\caption{Table For testing on YouTube video}
\begin{center}
\begin{tabular}{|c|c|c|}
\hline
\textbf{}&{\textbf{Observation by}}&\textbf{Observation by} \\
\textbf{}&{\textbf{human}}&\textbf{model} \\
%\cline{2-4} 

\textbf{YouTube Video Link} & \textbf{\textit{Total human}}&\textbf{\textit{Total warning}} \\
\textbf{} & \textbf{\textit{approached}}&\textbf{\textit{generated}} \\
\textbf{} & \textbf{\textit{}}&\textbf{\textit{(Correctly)}} \\
\hline
www.youtube.com/watch & 9 & 8 \\
?v=EXUQnLyc3yE\cite{youtube1} & & \\
\hline
www.youtube.com/watch & 10 & 9 \\
?v=SotdSIWmg7Q\cite{youtube2} & & \\
\hline
www.youtube.com/watch & 8 & 8 \\
?v=6NBwbKMyzEE\cite{youtube3} & & \\
\hline

%{$^{\mathrm{a}}$Table for model testing on YouTube videos.}
\end{tabular}
\label{tab1}
\end{center}
\end{table}

\section{Conclusion and Future Work} As technology is growing these days and as it is making the life of a human being easier and easier this is our little effort to make the life of visually impaired persons easier using the latest deep learning technologies. We have created our model in such a way that it can perform in real-time so that it can be really helpful. At the same time, our model should be more and more accurate as it will be assisting the blind person. That is why we are using optimal models which are neither too heavy nor too much inaccurate. In the future, we will try to find solutions for different environmental conditions and will try to make a better model which can work on any environmental condition and in real time.

\printbibliography
\end{document}